\def\year{2022}\relax
\documentclass[letterpaper]{article} 
\usepackage{aaai22}  
\usepackage{times}  
\usepackage{helvet}  
\usepackage{courier}  
\usepackage[hyphens]{url}  
\usepackage{graphicx} 
\usepackage{natbib}  
\usepackage{caption} 
\usepackage{booktabs}
\DeclareCaptionStyle{ruled}{labelfont=normalfont,labelsep=colon,strut=off} 
\usepackage{algorithm}
\usepackage{algorithmic}
\usepackage{amsmath}
\usepackage{amssymb}
\usepackage{amsfonts}
\usepackage{makecell}
\newcommand{\tabitem}{~~\llap{\textbullet}~~}
\usepackage{newfloat}
\usepackage{listings}
\floatstyle{ruled}
\newfloat{listing}{tb}{lst}{}
\floatname{listing}{Listing}
\newtheorem{theorem}{Theorem}
\newtheorem{prop}{Proposition}

\newtheorem{example}{Example}

\newcommand{\vapprox}{\mathrel{\scalebox{1}[2]{$\shortmid$}\mkern-3.1mu\raisebox{0.3ex}{$\approx$}}}
\title{Differentiable Fuzzy $\mathcal{ALC}$: \\ A Neural-Symbolic Representation Language for Symbol Grounding}
\author{
    Xuan Wu, Xinhao Zhu, Yizheng Zhao, Xinyu Dai
}
\affiliations{

    National Key Laboratory for Novel Software Technology,\\ Nanjing University, Nanjing, 210023, China
}

\usepackage{bibentry}

\begin{document}

\maketitle

\begin{abstract}
Neural-symbolic computing aims at integrating robust neural learning and sound symbolic reasoning into a single framework, so as to leverage the complementary strengths of both of these, seemingly unrelated (maybe even contradictory) AI paradigms. The central challenge in neural-symbolic computing is to unify the formulation of neural learning and symbolic reasoning into a single framework with common semantics, that is, to seek a joint representation between a neural model and a logical theory that can support the basic grounding learned by the neural model and also stick to the semantics of the logical theory. In this paper, we propose differentiable fuzzy $\mathcal{ALC}$ (DF-$\mathcal{ALC}$) for this role, as a neural-symbolic representation language with the desired semantics. DF-$\mathcal{ALC}$ unifies the description logic $\mathcal{ALC}$ and neural models for symbol grounding; in particular, it infuses an $\mathcal{ALC}$ knowledge base into neural models through differentiable concept and role embeddings. We define a hierarchical loss to the constraint that the grounding learned by neural models must be semantically consistent with $\mathcal{ALC}$ knowledge bases. And we find that capturing the semantics in grounding solely by maximizing satisfiability cannot revise grounding rationally. We further define a rule-based loss for DF adapting to symbol grounding problems. The experiment results show that DF-$\mathcal{ALC}$ with rule-based loss can improve the performance of image object detectors in an unsupervised learning way, even in low-resource situations.

\end{abstract}

\section{Introduction}
\label{sec:intro}
For decades, trends in the computational modeling of intelligent behavior have followed a recurring pattern, cycling between a primary focus on symbolic methods and sub-symbolic methods. The symbolic approach believes that the best way to nurture an AI is to feed it human-readable information related to what you think it needs to know to become capable of solving a particular human-level intelligence task. On the other hand, the sub-symbolic approach admits that human-based information formats are not always the best fit for an AI, and encourages feeding raw data into the AI so that it can construct its own implicit knowledge and have its own way to interpret the meaning of the data. 
To find an in-between solution of symbolic and sub-symbolic methods, neural-symbolic computing~\citep{hammer2007perspectives,garcez2015neural,DBLP:journals/ai/DiligentiGS17,DBLP:journals/flap/GarcezGLSST19,DBLP:journals/jwe/HarmelenT19,garcez2020neurosymbolic,DBLP:conf/iclr/LampleC20,kahneman2020aaai20,marcus2020next} arises, which aims to enable AI agents' cognitive ability, containing interpreting (reasoning) the raw data according to the given knowledge in a human-readable and machine-readable way.

One main challenge of neural-symbolic computing is to interpret the meaning of objects (e.g. results of sub-symbolic methods) with human-readable semantics (e.g. structured symbols), which is to ground symbols to these objects, also known as symbol grounding problem~\citep{harnad1990symbol,cangelosi2011solutions,coradeschi2013short}.  
Sub-symbolic methods may be one way to ground concrete instances in the capacity to categorize them ~\citep{harnad1993symbol} but fail to capture the inherent logical relations between symbols.
Many artificial intelligence tasks can be reduced to the symbol grounding problem, as shown in Table.~\ref{intro:1}. Neural-symbolic methods vary in the way that they use two kinds of knowledge: propositional (know-what/declarative) knowledge and non-propositional (know-how/procedural) knowledge~\citep{goldstein1977artificial,charest2006ontology}, most of which~\citep{jiang2021lnn,liang2017neural,zhang2020learning,riegel2020logical} only considered procedural knowledge. 

But it is worth noting that procedural and declarative knowledge should not be used in the same way, which is a challenge to integrate the sub-symbolic and symbolic systems.
For classification, the procedural knowledge is to assign similar instances to the same class, i.e., $\mathtt{isKindOf}(x_1,A)\land \mathtt{SimilarEntity}(x_1,x_2)\implies \mathtt{isKindOf}(x_2,A)$.  The $\mathtt{SimilarEntity}$ predicate is learned from the feature of the entity, under the hypothesis that the similar structure or context the pair of entities have, the higher probability that they are the same entity, which is similar to the distributional hypothesis, which is not extracted in a human-readable way. So the result is monitored by the distribution of data, and the knowledge should be used in a soft way.  While declarative knowledge  (e.g. $\mathtt{armChair}(x_1)\implies \mathtt{has}(x_1,\mathtt{ChairArm})\land \mathtt{has}(x_1,\mathtt{Chair})$) should be compulsory for AI system to admit to human society. So the encoding of declarative knowledge should not lose the logical semantics.
Compared to past works, we consider two symbol grounding ways between sub-symbolic methods and two kinds of knowledge separately.

\begin{table*}[ht]
    \resizebox{\textwidth}{!}{
    \begin{tabular}{@{}ccccc@{}}
    \toprule
      Tasks & Raw Data & Symbol System & Descriptions & Neural-symbolic Works \\ \midrule
      
       \makecell{Semantic Image\\ Interpretation} & images & \makecell{\tabitem $\forall xy(\mathtt{Saddle}(x)\land\mathtt{partOf}(x,y)\rightarrow \mathtt{Motorcycle}(y)\lor \mathtt{Bicycle}(y))$\\} & \makecell{
           \tabitem Knowledge depicts scene, and \\logical relationships between objects \\
           \tabitem Ground scene concepts (propositions) to images,\\ object concepts region in images
       } & \citep{donadello2017logic,arvor2019ontologies,oltramari2020neuro} \\\midrule

   	\makecell{ Named-entity\\ Recognition} & \makecell{natural language\\ utterance} & 
     \makecell{ \tabitem $\forall x((x=``Organization\ of\  Java'')$\\$\implies \mathtt{NNP}(chunk(x,1))\land \mathtt{IN}(chunk(x,2))\land\mathtt{NNP}(chunk(x,3))$, \\
      \tabitem $\forall x(\mathtt{Begin}(tag(x,0))$\\$\implies \mathtt{Interior}(tag(x,1)) \lor \mathtt{End}(tag(x,1)))$,\\
      \tabitem $\forall x(\mathtt{isMention}(x)\land\mathtt{isEntity}(x,European\_Union)$\\$\implies \mathtt{isEntity}(x,EU))$
     } &
     \makecell{\tabitem Knowledge about part-of-speech tagging sequence, \\ span recognition, type associations \\
     \tabitem Ground named-entity to entity mentions
     }
     & \citep{seyler2018study,hu2016harnessing,torisawa2007exploiting}\\\midrule

     Schema Matching & \makecell{two knowledge bases;\\ natural language question with a knowledge base} & \tabitem $ \mathtt{SameEntity}(x_1,x_2)\land\mathtt{isKindOf}(x_1,A)\implies\mathtt{isKindOf}(x_2,A)$ & \makecell{
           \tabitem Definition about schema matching \\
           \tabitem Ground the proposition $\mathtt{SameEntity}$ to \\ entities from different resource with the same semantics
       } &\cite{jiang2021lnn,liang2017neural,unal2006using}\\\midrule
       
       Link Prediction & \makecell{graph-structured data, \\e.g. knowledge graph, \\ biological network,\\ social network} &   \makecell{ \tabitem $\forall x,y,s,r(\mathtt{Node}(x)\land \mathtt{Node}(y)\land\mathtt{Node}(s)\land\mathtt{Relation}(r)$ \\$\land \mathtt{Link}(r,x,y)\land\mathtt{Similar}(x,s)\implies \mathtt{Link}(r,s,y))$
    ,  \\ \tabitem $\forall x,y,s,r(\mathtt{Node}(x)\land \mathtt{Node}(y)\land\mathtt{Relation}(s)\land\mathtt{Relation}(r)\land $\\ $\mathtt{Link}(r,x,y)\land\mathtt{Similar}(r,s)\implies \mathtt{Link}(s,x,y))$}& \makecell{
           \tabitem Intuition for link prediction \\
           \tabitem Ground the linked entities to \\ entities with the similar semantics
       }&     \cite{zhang2020learning,trouillon2016complex}\\\midrule
     
       \bottomrule
       
    \end{tabular}
    }
    \caption{AI Tasks aims to ground symbol system (relevant concepts) in raw data. }
    \label{intro:1}
\end{table*}

Another bottleneck of symbol grounding is knowledge representation. 
Existing neural-symbolic works ignore the importance of choosing a logic language that can ``optimal'' balance the expressive power with computational complexity, as using declarative and knowledge requires the full logic semantics to be maintained in the neural-symbolic representation. Knowledge graph (KG) embedding methods~\citep{lamb2020graph,xie2019embedding,zhang2021neural}, which are generally based on resource description framework (RDF), achieve great success in many applications, especially in link prediction, while deductive/abductive reasoning in the graph is inefficient (as inductive reasoning should be done first) and not fully explainable. In other words, RDF-based KG has not enough expressivity to satisfy the requirement of neural-symbolic computing.
While the expressive power of first-order logic (FOL) is too strong, accompanying undecidability. FOL-based neural-symbolic methods are mostly used in experimental tasks~\citep{riegel2020logical,badreddine2022logic,DBLP:journals/flap/GarcezGLSST19,garcez2019neural}, and ignore analyzing in what situations the semantics of FOL axiom can be lost in computing. Description logics (DLs) as a family of decidable segments of FOL, balance the expressive power and complexity. Besides, DLs as the core of ontology web language (OWL) are used in many applications, the deductive complexity of which is well-studied. 

To adapt description logic to a neuro-symbolic paradigm, we develop a differentiable fuzzy language based on $\mathcal{ALC}$, which can depict knowledge under unstable situations and band between sub-symbolic system and $\mathcal{ALC}$ ontology without losing the logical semantics.
 Behind the choice of $\mathcal{ALC}$ is a series of deliberations: 1) currently, $\mathcal{EL}++$ is the logic language with the highest expressive power with fully captured logic semantics in existing work~\citep{kulmanov2019embeddings}, but when considering the negation operator, all current knowledge embedding models cannot ensure the semantics are not lost; $\mathcal{ALC}$ is with high expressive power as it introduces negation beyond $\mathcal{EL}$, but there is no work banding the logic semantics of $\mathcal{ALC}$ ontology with sub-symbolic methods; fuzzy $\mathcal{ALC}$ based on G\"{o}del semantics is decidable~\citep{baader2017decidability,borgwardt2014decidable,borgwardt2014godel} without finite model property, which means that traditional tableau-based reasoning algorithm for infinitely valued fuzzy $\mathcal{ALC}$ is not sound.

In this work, we propose differentiable fuzzy $\mathcal{ALC}$ (DF-$\mathcal{ALC}$) for symbol grounding, based on infinitely valued fuzzy $\mathcal{ALC}$. Different from the existing differentiable fuzzy language~\citep{van2022analyzing,badreddine2022logic}, we consider knowledge represented in a language that can contain more logical information in neuro-symbolic computing, and introduce two effective interpreting ways when given declarative/procedural knowledge.

The key contributions of this work can be summarized as follows:
\begin{itemize}
    \item We present a representation language DF-$\mathcal{ALC}$ which facilitates a sound and complete mechanism to revise the probabilistic semantics by a neural model according to a consistent $\mathcal{ALC}$ ontology. This makes us the first to combine neuro-symbolic computing with description logic.
    \item Rather than ground by maximizing satisfiability. We designed rule-based loss, which helps fuzzy description logic adapt to revise grounding rationally, and improves the performance in semantic image interpretation.  
     \item Experiments show that DF-$\mathcal{ALC}$ can keep the reliable component of the perceptual grounding. Meanwhile, unknown situations are few to affect grounding, this further demonstrates that the semantics of DF-$\mathcal{ALC}$ are solid in terms of crisp $\mathcal{ALC}$ under OWA. 
    \item We are the first to enable $\mathcal{ALC}$ ontology to be compatible with sub-symbolic methods while not losing the logical semantics. The source code, alongside the experimental settings, is publicly accessible at \url{https://anonymous.4open.science/r/DF-ALC}.
\end{itemize}

\section{Related Works}
Our work is an intersection branch of ontology representation learning and neuro-symbolic computing. And we give a feasible way out for a symbol grounding problem --- semantic image interpretation.

\subsection{Neuro-symbolic Computing}
Neural-symbolic computing aims at computing with both learning and reasoning abilities, to step towards the combination of symbolic and sub-symbolic systems. Current learning ability relies largely on differentiable programming to draw conclusions from observations and apply them, while current reasoning ability relies largely on logical programming to give conclusions inferred from premises and rules through deductive reasoning, give rules according to observations comprising premises and conclusions through inductive reasoning, and give premises that can interpret conclusions according to rules through abductive reasoning. So it comes with challenges in the integration and representation of these two kinds of programming paradigms. 
From the perspective of integration, research works differ in logical techniques that are mainly consumed. Neural-symbolic inductive logical programming~\citep{wang2013programming,buhmann2016dl,yang2017differentiable,evans2018learning,sen2022neuro} and statistical relational learning (e.g. Markov logic network~\citep{richardson2006markov}, probabilistic soft logic~\citep{bach2017hinge}) works seek to learn probabilistic logical rules from observations. This requires learning model parameters in a continuous space and the structure in a discrete space. SATNet~\cite{wang2019satnet} learns rules from labeled data by transforming the learning problem as SAT problem\footnote{this is the satisfiability (SAT) problem which aims to determine whether there exists an interpretation that satisfies a given formula}. 
To combine the ability of deductive reasoning, the first line of research learn to reason by modeling the inference procedure using neural networks or replacing logical computations with differentiable functions~\citep{towell1994knowledge,holldobler1999approximating,rocktaschel2016learning,rocktaschel2017end,diligenti2017semantic,ebrahimi2021capabilities}. But this neglects factual knowledge which bridges the physical world and the conceptual world, so the second line of research aims to find an interpretation (grounding) that satisfies theories which can be a mapping between these two worlds by encoding the satisfiability of theories in the loss function~\citep{badreddine2022logic,serafini2016logic,riegel2020logical,topan2021techniques,van2022analyzing}. The notable work Logical Tensor Network (LTN)~\cite{badreddine2022logic} uses neural networks to represent the fuzzy function and predicates of theories, which is learned from labeled data. To solve the symbol grounding problem, LTN learns the interpretation with trained parameters that can maximize the satisfiability of theories.
But these works cannot find explanations of observations according to theories, so abductive learning-based neural-symbolic works are proposed to use the explanations getting through abductive reasoning to promote the interpretability of the computing~\citep{zhou2019abductive,huang2020semi,tsamoura2021neural,cai2021abductive}. 
 From the view of representation, some works are based on classical logic --- propositional logic~\citep{towell1994knowledge,zhou2019abductive,tsamoura2021neural,cai2021abductive}, description logic~\citep{buhmann2016dl,eberhart2019completion,ebrahimi2021capabilities}, or first-order logic~\citep{holldobler1999approximating,wang2013programming,rocktaschel2016learning,serafini2016logic,rocktaschel2017end,yang2017differentiable,evans2018learning,sen2022neuro}, others are based on non-classical logic, such as fuzzy logic~\citep{diligenti2017semantic,riegel2020logical,van2022analyzing}, or probabilistic logic~\citep{wang2013programming,manhaeve2018deepproblog}. 
Our work tries to combine the deductive reasoning ability in a novel way, which follows the work of LTN. As stated in the introduction, when using declarative knowledge, data bias should be ignored, so different from works that only maximize the satisfiability of the knowledge base, we give a compulsory way to inject knowledge.

\subsection{Ontology Representation Learning}
The main motivation for embedding OWL ontology into vector space is to transfer the knowledge to vectors so it can be used directly in knowledge-requiring downstream tasks. Two kinds of methods exist with different input requirements. The first kind focuses on coupling the meta-data of an OWL ontology into an efficient graph, and then uses the generated corpus based on the graph as the input to the existing representation learning methods~\citep{smaili2018onto2vec,smaili2019opa2vec,chen2021owl2vec}. The second kind focuses on modeling the logical semantics of an OWL ontology. EL2Vec~\citep{kulmanov2019embeddings} approximates geometric models for $\mathcal{EL}++$ ontologies and has achieved an interpretable embedding for GO. E2R~\citep{garg2019quantum} can model the logical operators of intersection, union, negation, and universal quantifier, but fails to capture the distributive law. In all, there often comes a loss of the semantics of an OWL ontology in the transformation of most embedding methods~\citep{smaili2018onto2vec,smaili2019opa2vec,chen2021owl2vec}. Though the geometric construction method~\citep{kulmanov2019embeddings} can preserve the logical semantics well in $\mathcal{EL}++$, the embedding may bring unexpected knowledge (unknown becomes true) because of the closed world assumption (CWA), and $\mathcal{EL++}$ is not expressive as $\mathcal{ALC}$ studies in this work.

\subsection{Semantic Image Interpretation}
A symbol grounding application of our work is semantic image interpretation (SII) ~\cite{hudelot2005symbol,neumann2008scene,krishna2017visual,donadello2017logic}, which aims to generate a structured and human-readable description of the content of images. Current successful SII  researches~\cite{donadello2017logic,arvor2019ontologies} rely on background knowledge of the images. LTN~\cite{donadello2017logic} models predicates and functions as neural networks and learns the representation through maximizing the satisfiability in a supervised way. The main struggle of these neural-symbolic works in leveraging logical knowledge to adapt to the symbol grounding problem is that the revision signal cannot be properly conveyed.
Some early works~\cite{hudelot2008fuzzy,dasiopoulou2009applying} revise image semantics solely with fuzzy description logic, the idea of which can be summarized as selecting plausible image scene descriptions (assertions) ignoring disjoint cases, and then handle the inconsistency by removing assertions using methods e.g. reversed tableaux expansion procedure. Though these fuzzy description logic-based methods can remain as much reliable parts of perceptual grounding as possible, these works achieve most of the expected revisions and are specific to the ontology designed for image scene classification, which lacks generalization ability.

\section{Preliminaries}

\subsection{The Description Logic $\mathcal{ALC}$}

Let $N_C$ and $N_R$ be pairwise disjoint and countably infinite sets of \textit{concept names} and \textit{role names}, respectively. $\mathcal{ALC}$\textit{-concepts} are inductively constructed based on the following syntax rule:
\[C, D \rightarrow\top | \bot | A | \neg C | C\sqcap D | C\sqcup D | \exists r.C | \forall r.C,\]
where $A\in N_C$, $r\in N_R$, and $C$ and $D$ range over concepts. A concept of the form $A\in N_C$ is called \textit{atomic}, otherwise, it is \textit{compound}. An ontology $\mathcal{O}$ consists of a TBox and an ABox.
An $\mathcal{ALC}$\textit{-TBox} $\mathcal{T}$ is a finite set of axioms of the form:
\begin{itemize}
    \item $C\sqsubseteq D$\textit{~(concept inclusion)}, and
    \item $C\equiv D$\textit{~(concept equivalence)},
\end{itemize}
where $C$ and $D$ are concepts. The \emph{disjointness} between $C,D$ is $C\sqcap D\sqsubseteq\bot$. We use the axiom $C\equiv D$ as abbreviation for $C\sqsubseteq D$ and $D\sqsubseteq C$.

Let $N_I$ be disjoint and countably infinite sets of \emph{individual names}, while an $\mathcal{ALC}\textit{-ABox}$ $\mathcal{A}$ is a finite set of crisp assertions of the form:
\begin{itemize}
    \item $a:C$\textit{~(concept assertion)} , and
    \item $(a,b):r$ \textit{~(role assertion)},
\end{itemize}
 where $C$ is a concept, $r$ is a role name, and $a,b$ are individuals from $N_I$.

An $\mathcal{ALC}$ ontology is comprised of an $\mathcal{T}$ and $\mathcal{A}$, denoted as $\mathcal{O}=\langle\mathcal{T},\mathcal{A}\rangle$. 
The \emph{signature} of $\mathcal{O}$ is $sig(\mathcal{O})=N_C\cup N_R\cup N_I$.

The semantics of $\mathcal{O}$ is defined in terms of an \textit{interpretation} $\mathcal{I}=\langle\Delta^{\mathcal{I}}, \cdot^{\mathcal{I}}\rangle$, where $\Delta^{\mathcal{I}}$ denotes the \textit{domain of the interpretation} (a non-empty \textit{crisp set}), and $\cdot^{\mathcal{I}}$ denotes the \textit{interpretation function}, which assigns to every concept name $A\in N_C$ a set $A^{\mathcal{I}}\subseteq \Delta^{\mathcal{I}}$, and to every role name $r\in N_R$ a binary relation $r^{\mathcal{I}}\subseteq \Delta^{\mathcal{I}}\times \Delta^{\mathcal{I}}$. The interpretation function $\cdot^{\mathcal{I}}$ is inductively extended to concepts as follows:
    \[\top^{\mathcal{I}}  = \Delta^{\mathcal{I}}, \bot^{\mathcal{I}} = \emptyset, (\neg C)^{\mathcal{I}}=\Delta^{\mathcal{I}}\backslash C^{\mathcal{I}}, \]
     \[ (C\sqcap D)^{\mathcal{I}}  = C^{\mathcal{I}}\cap D^{\mathcal{I}}, (C\sqcup D)^{\mathcal{I}}  = C^{\mathcal{I}}\cup D^{\mathcal{I}},\]
    \[ (\exists r.C)^{\mathcal{I}}  = \{a \in \Delta^{\mathcal{I}}~\vert~\exists b.
(a,b) \in r^{\mathcal{I}} \wedge b \in C^{\mathcal{I}}\},\]
 \[(\forall r.C)^{\mathcal{I}}  = \{a \in \Delta^{\mathcal{I}}~\vert~\forall b.
(a,b) \in r^{\mathcal{I}} \rightarrow b \in C^{\mathcal{I}}\}.\]

Let~$\mathcal{I}$ be an interpretation. A concept inclusion ${C\sqsubseteq D}$ is \emph{true} in~$\mathcal{I}$ iff $C^{\mathcal{I}}\subseteq D^{\mathcal{I}}$. A concept assertion $a : A$ is \emph{true} in $\mathcal{I}$ iff $a^{\mathcal{I}}\in C^{\mathcal{I}}$. A role assertion $(a, b) : r$ is \emph{true} in $\mathcal{I}$ iff $(a^{\mathcal{I}}, b^{\mathcal{I}})\in r^{\mathcal{I}}$. $\mathcal{I}$ is a \emph{model} of an ontology $\mathcal{O}$, write $\mathcal{I}\models\mathcal{O}$, iff every axiom in $\mathcal{O}$ is \emph{true} in~$\mathcal{I}$. An axiom $\beta$ is entailed by an ontology $\mathcal{O}$, write $\mathcal{O}\models\beta$, iff $\beta$ is true in every model $\mathcal{I}$ of $\mathcal{O}$. An ontology $\mathcal{V}$ is entailed by another ontology $\mathcal{O}$, write $\mathcal{O}\models\mathcal{V}$, iff every model of $\mathcal{V}$ is also a model of $\mathcal{O}$. An ontology $\mathcal{O}$ is \emph{consistent (true)} if there exists a model~$\mathcal{I}$ of~$\mathcal{O}$.
A concept $C$ is satisfiable w.r.t. ~$\mathcal{O}$ if there exists a model~$\mathcal{I}$ of~$\mathcal{O}$ and some $d\in \Delta^\mathcal{I}$ with $d\in C^\mathcal{I}$. 
A concept assertion ${a: C}$ is satisfiable in~$\mathcal{I}$ iff $a^{\mathcal{I}}\in C^{\mathcal{I}}$. A role assertion ${(a,b):r}$ is satisfiable in~$\mathcal{I}$ iff $(a^{\mathcal{I}}, b^{\mathcal{I}})\in r^{\mathcal{I}}$.  

Other basic reasoning problems are polynomial-time reducible to the satisfiability problem.
A concept inclusion ${C\sqsubseteq D}$ is true in~$\mathcal{I}$ iff the concept $C\sqcap\neg D$ is unsatisfiable in~$\mathcal{I}$. The retrieval problem of computing the instantiation of concept $C$ is polynomial-time reducible to that of checking the satisfiability of ${a: C}$.

Under the interpretation $\mathcal{I}$, concepts and roles are mapped into crisp sets in $\Delta^{\mathcal{I}}$, so the vagueness cannot be modeled. 
\begin{figure*}[ht]
    \centering
    \includegraphics[width=\linewidth]{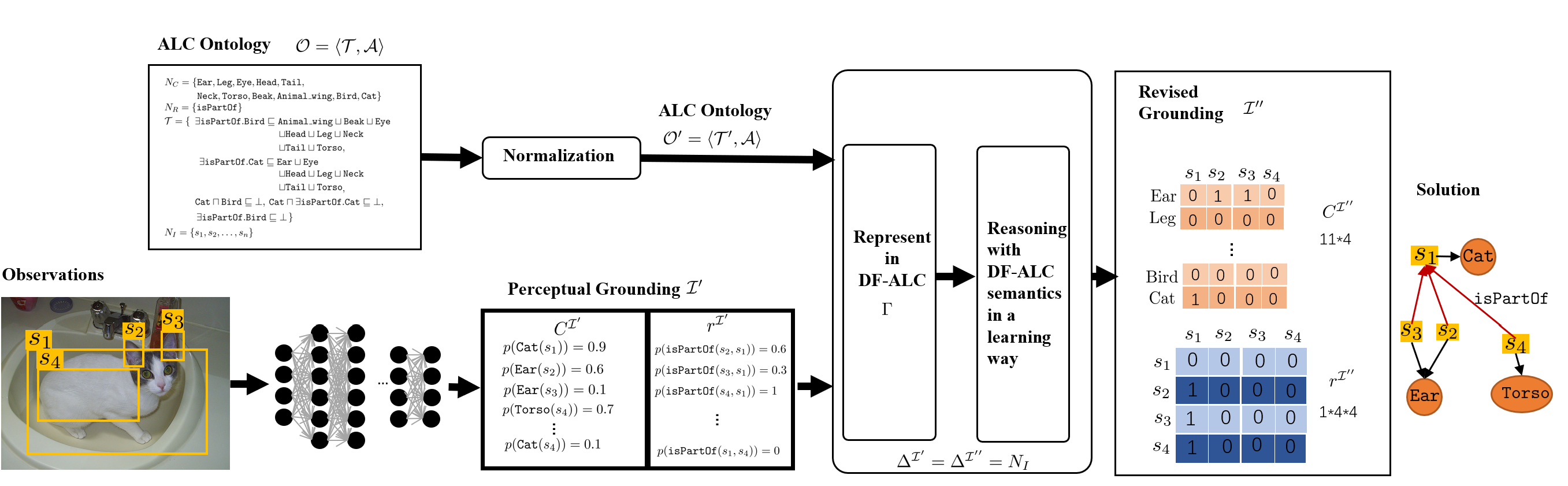}
    \caption{An ontology-based semantic image interpretation example utilizing DF-$\mathcal{ALC}$. Ontology gives background knowledge about the cat and bird. The neural model does the low-level analysis for the image, which gives the wrong grounding for the image.}
    \label{fig:df-alc}
\end{figure*}

\subsection{Fuzzy $\mathcal{ALC}$}
Fuzzy set theory and fuzzy logic were proposed by Zadeh~\citep{zadeh1996fuzzy} to manage imprecise and vague knowledge.
Based on fuzzy set theory~\citep{zadeh1996fuzzy}, a \emph{fuzzy set} $X$ w.r.t. an universe is characterized by a \emph{membership function} $\mu_X:U\rightarrow[0,1]$. Each element $u\in U$ is assigned with an $X$-membership degree $\mu_X(u)$. 
In fuzzy logic, $\mu_X(u)$ is the \emph{truth-value} of the statement `$u$ is $X$'.

Fuzzy $\mathcal{ALC}$ retains the same syntax with $\mathcal{ALC}$, only semantics changes.
Here, we follow fuzzy $\mathcal{ALC}$ proposed in ~\cite{straccia2001reasoning}, which is based on G\"{o}del logic~\citep{dummett1959wittgenstein}. 

A fuzzy interpretation (also called \emph{grounding} here) $\mathcal{I}$ consists of a non-empty domain $\Delta^{\mathcal{I}}$ and an \emph{interpretation function} $\cdot^{\mathcal{I}}$ defined as: (1) an individual $a$ is interpreted by $\mathcal{I}$ as an element $a^{\mathcal{I}}\in\Delta^{\mathcal{I}}$, and; (2) a concept $C$ is interpreted by $\mathcal{I}$ as a fuzzy set $C^{\mathcal{I}}:\Delta^{\mathcal{I}}\rightarrow[0,1]$, and; (3) a role $r$ is interpreted by $\mathcal{I}$ as a fuzzy set $r^{\mathcal{I}}:\Delta^{\mathcal{I}}\times\Delta^{\mathcal{I}}\rightarrow[0,1]$.

The fuzzy interpretation function $\cdot^{\mathcal{I}}$ is inductively extended to concepts as follows, for all $a\in\Delta^{\mathcal{I}}$:
\begin{equation}
\label{e:1}
    \top^{\mathcal{I}}(a)  = 1, \bot^{\mathcal{I}}(a)=0, (\neg C)^{\mathcal{I}}(a)=1- C^{\mathcal{I}}(a),
\end{equation}
\begin{equation}
\label{e:2}
    (C\sqcap D)^{\mathcal{I}}(a)  = \mathrm{min}\{C^{\mathcal{I}}(a), D^{\mathcal{I}}(a)\},
\end{equation}
\begin{equation}
\label{e:3}
    (C\sqcup D)^{\mathcal{I}}(a)  = \mathrm{max}\{C^{\mathcal{I}}(a), D^{\mathcal{I}}(a)\},
\end{equation}
\begin{equation}
\label{e:4}
    (\exists r.C)^{\mathcal{I}}(a)  = \mathrm{sup}_{b\in\Delta^\mathcal{I}}\{\mathrm{min}\{r^\mathcal{I}(a,b),C^\mathcal{I}(b)\}\},
\end{equation}
\begin{equation}
\label{e:5}
    (\forall r.C)^{\mathcal{I}}(a)  = \mathrm{inf}_{b\in\Delta^\mathcal{I}}\{\mathrm{max}\{1-r^\mathcal{I}(a,b),C^\mathcal{I}(b)\}\}.
\end{equation}

A fuzzy $\mathcal{ALC}$ TBox is a finite set of \emph{fuzzy inclusion} of the form ${C\sqsubseteq D}$. ${C\sqsubseteq D}$ is true (i.e., truth-value is 1) in $\mathcal{I}$ (or we say $\mathcal{I}$ satisfies $C\sqsubseteq D$) iff,
 \begin{equation}
    \forall a\in\Delta^{\mathcal{I}},\ C^{\mathcal{I}}(a)\leq D^{\mathcal{I}}(a)
\end{equation}
We say that two concepts $C$ and $D$ are \emph{fuzzy equivalent} ($C\cong D$) when $C^\mathcal{I}(a)=D^\mathcal{I}(a)$ for all $a\in\Delta^{I}$.

A fuzzy $\mathcal{ALC}$ ABox is a finite set of \emph{fuzzy assertion} of the form $(a:C)\bowtie n$ or $(\langle a,b\rangle:r)\bowtie n$, where $\bowtie$ stands for $\geq,>,\leq,<$, and $n\in[0,1]$ is the \emph{truth value}. Formally, a fuzzy interpretation  $\mathcal{I}$ satisfies a fuzzy assertion $(a:C)\bowtie n$ (resp. $(\langle a,b\rangle:r)\bowtie n$) iff $C^{\mathcal{I}}(a^{\mathcal{I}}) \bowtie n$ (resp. $r^\mathcal{I}(a^{\mathcal{I}},b^{\mathcal{I}}) \bowtie n$). For simplicity, we write a fuzzy assertion as $\varphi=\{\phi\bowtie n\}$, where $\phi$ is $(a:C)$ or $(\langle a,b\rangle:r)$.

A fuzzy interpretation $\mathcal{I}$ is a \emph{model} of a fuzzy ontology $\mathcal{O}$, write $\mathcal{I}\vapprox\mathcal{O}$, iff $\mathcal{I}$ satisfies each axiom in $\mathcal{O}$. A fuzzy ontology $\mathcal{O}$ \emph{fuzzy entails} a fuzzy assertion $\varphi$, write $\mathcal{O}\vapprox\varphi$ iff every model of $\mathcal{O}$ also satisfies $\varphi$.

A \emph{crisp} $\mathcal{ALC}$ ontology is a specialism of fuzzy $\mathcal{ALC}$ ontology, and can easily be extended to a fuzzy ontology by assigning truth value 1 to assertions.

\subsection{Ontology-based Semantic Image Interpretation}
Let $S=\{s_1,...,s_n\}$ be a set of segments (a segment is a set of contiguous pixels) returned by a low-level analysis (e.g. object detection) of picture $\mathcal{P}$.
Given an ontology $\mathcal{O}$, the semantic image interpretation task can be formed as labeling picture $\mathcal{P}$ with an interpretation $\mathcal{I}$ defined in the domain $S$, which maps each segment $s\in S$ to a set of values $\{C^{\mathcal{I}}(s) ~\vert~ C\text{ is any concept in }\mathcal{O}\}$.

\section{Differentiable Fuzzy $\mathcal{ALC}$}
\label{sec:3}
DF-$\mathcal{ALC}$ is an extension of fuzzy $\mathcal{ALC}$. The semantics of ontology represented in DF-$\mathcal{ALC}$ can be infused into symbol grounding in a continuous space. To solve a symbol grounding problem, such as the semantic image interpretation problem shown in Figure~\ref{fig:df-alc}, where the concept symbols are about cat, bird and their components and the $\texttt{isPartOf}(s_2,s_1)$ relation symbol denotes image segment $s_2$ is a part of $s_1$. When the neural model is poorly trained (e.g. in a low-resource situation), DF-$\mathcal{ALC}$-represented knowledge helps to revise the perceptual grounding. The $\mathcal{ALC}$ ontology $\mathcal{O}=\langle\mathcal{T},\mathcal{A}\rangle$ is one of the inputs of our model, which contains the inherent relations between symbols. Another input, the perceptual grounding $\mathcal{I}'$ contains valuable information for grounding but is not a fuzzy model of fuzzy extended $\mathcal{O}$, so it is not reasonable. To get the grounding that can retain information of $\mathcal{I}'$ as much as possible, and is a model of $\mathcal{O}$, $\mathcal{T}$ is normalized to keep axioms in the standard format, and the normalized ontology $\mathcal{O}'$ is reformulated into DF-$\mathcal{ALC}$ representation. Revising perceptual grounding is realized by maximizing the satisfiability of the reformulated DF-$\mathcal{ALC}$ ontology. When the revision is finished, the revised grounding is a model of $\mathcal{O}$.

\subsection{Normalization}
\begin{figure}[ht]
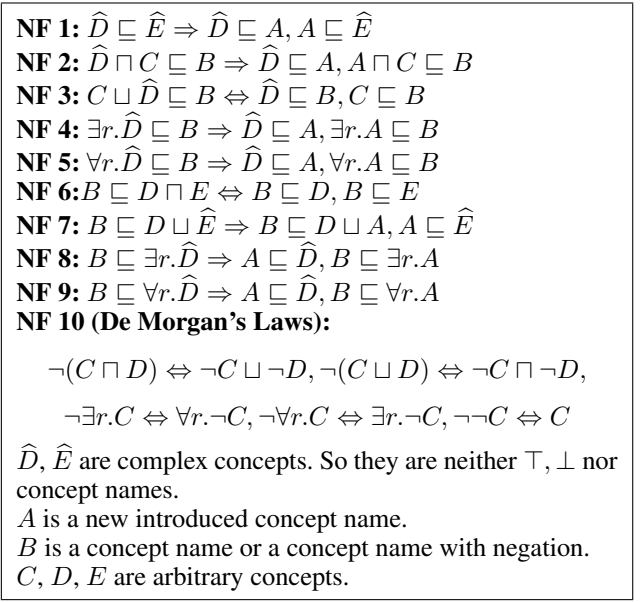

\fbox{
\begin{minipage}{0.95\columnwidth}
    \textbf{NF 1:}
    $\widehat{D}\sqsubseteq \widehat{E} \Rightarrow \widehat{D}\sqsubseteq A, A\sqsubseteq \widehat{E}$

    \textbf{NF 2:}
    $\widehat{D}\sqcap C \sqsubseteq B \Rightarrow \widehat{D}\sqsubseteq A, A\sqcap C\sqsubseteq B$
    
    \textbf{NF 3:}
    $C\sqcup \widehat{D}\sqsubseteq B \Leftrightarrow \widehat{D}\sqsubseteq B, C\sqsubseteq B$
    
    \textbf{NF 4:}
    $\exists r.\widehat{D}\sqsubseteq B\Rightarrow \widehat{D}\sqsubseteq A, \exists r.A\sqsubseteq B$
    
    \textbf{NF 5:}
    $\forall r.\widehat{D}\sqsubseteq B\Rightarrow\widehat{D}\sqsubseteq A, \forall r.A\sqsubseteq B$
    
    \textbf{NF 6:}$B\sqsubseteq D\sqcap E\Leftrightarrow B\sqsubseteq D, B\sqsubseteq E$

    \textbf{NF 7:}
    $B\sqsubseteq D\sqcup \widehat{E}\Rightarrow B\sqsubseteq D\sqcup A, A\sqsubseteq \widehat{E}$

    \textbf{NF 8:}
    $B\sqsubseteq\exists r.\widehat{D}\Rightarrow A\sqsubseteq\widehat{D}, B\sqsubseteq\exists r.A$
    
    \textbf{NF 9:}
    $B\sqsubseteq\forall r.\widehat{D}\Rightarrow A\sqsubseteq\widehat{D}, B\sqsubseteq \forall r.A$
    
    \textbf{NF 10 (De Morgan's Laws):}
    \[\neg(C\sqcap D)\Leftrightarrow \neg C\sqcup\neg D, \neg(C\sqcup D)\Leftrightarrow \neg C\sqcap \neg D, \]
    \[\neg\exists r.C\Leftrightarrow \forall r.\neg C, \neg\forall r.C\Leftrightarrow\exists r.\neg C, \neg\neg C\Leftrightarrow C\]

    $\widehat{D}$, $\widehat{E}$ are complex concepts. So they are neither $\top,\bot$ nor concept names. \\
    $A$ is a new introduced concept name.  \\
    $B$ is a concept name or a concept name with negation. \\
    $C$, $D$, $E$ are arbitrary concepts.
\end{minipage}
}

\caption{Normalization rules for $\mathcal{ALC}$}
\label{fig:alc-normalization}
\end{figure}

Given an $\mathcal{ALC}$ ontology $\mathcal{O}=\langle\mathcal{T},\mathcal{A}\rangle$, concepts in $\mathcal{T}$ are transformed into negation normal forms using De Morgan's Laws until all concepts have no indirect negation. Then we recursively apply NF1-9 in Figure~\ref{fig:alc-normalization} until all the axioms are in the forms in Figure~\ref{fig:normalized-forms}.

Any axioms in an $\mathcal{ALC}$ ontology can be transformed into the normal forms in Figure~\ref{fig:normalized-forms}, using the rules in Figure~\ref{fig:alc-normalization}. For the concept equivalence axiom $C\equiv D\Leftrightarrow C\sqsubseteq D, D\sqsubseteq C$, two inclusions should use the same introduced concept names set. The introduced concepts should also not interfere with the semantics of the ontology, so a logical reasoner is used here to add assertions about introduced concept names to the ABox. Theorem~\ref{theorem:normalization} ensures that the interpretation of the normalized ontology is semantically equivalent to the interpretation of $\mathcal{O}$.

\begin{figure}[ht]
    \centering
    \fbox{
    \begin{minipage}{0.95\columnwidth}
        \textbf{Form 1:} $C\sqsubseteq B$ \\
        \textbf{Form 2:} $C_1\sqcap C_2\sqsubseteq B$ \qquad\qquad \textbf{Form 3:} $B\sqsubseteq C_1\sqcup C_2$\\
          \textbf{Form 4:} $C\sqsubseteq\exists r.B$ \qquad\qquad\quad\;  \textbf{Form 5}: $C\sqsubseteq\forall r.B$\\
          \textbf{Form 6}: $\exists r.B\sqsubseteq C$ \qquad\qquad\quad\;  \textbf{Form 7:} $\forall r.B\sqsubseteq C$\\
        
        $C$, $B$, $C_1$, $C_2$ are concept names or concept names with negation. \\
        $r$ is a role name.
        
    \end{minipage}
    }
    \caption{Normalized forms of $\mathcal{ALC}$ ontologies for DF-$\mathcal{ALC}$}
    \label{fig:normalized-forms}
\end{figure}

\begin{theorem}
\label{theorem:normalization}
For any $\mathcal{ALC}$ ontology $\mathcal{O}$, one can construct in polynomial time a normalized $\mathcal{ALC}$-ontology $\mathcal{O}$' of polynomial size in $|\mathcal{O}|$ using the normalization described above such that (i) for every model $\mathcal{I}$ of \ $\mathcal{O}$, there exists a model $\mathcal{J}$ of \ $\mathcal{O}'$ such that $\mathcal{I}$ is semantically equivalent to $\mathcal{J}$ in \ $sig(O)$, denoted as $\mathcal{I}\sim_{sig(O)}\mathcal{J}$, and (ii) for every model $\mathcal{J}$ of $\mathcal{O}'$ there exists a model $\mathcal{I}$ of $\mathcal{O}$ such that $\mathcal{I}\sim_{sig(O)}\mathcal{J}$. 
\end{theorem}

After normalization, we can see that formula in the form of Figure~\ref{fig:normalized-forms} has at most one logical operation except the subclass operator, so any normalized $\mathcal{ALC}$ ontology in Figure ~\ref{fig:alc-normalization} can be efficiently used as the input to a neural network.

\subsection{Syntax}
The syntax of DF-$\mathcal{ALC}$ mainly differs from fuzzy $\mathcal{ALC}$ in the definition of fuzzy assertion. For each fuzzy assertion $\varphi=\{\phi\bowtie n\}$, we set the $\bowtie$ be $=$ by considering two assertions of the from $\phi\geq n$ and $\phi\leq n$.
Then we introduce how to transform the given $\mathcal{ALC}$ ontology $\mathcal{O}=\langle\mathcal{T},\mathcal{A}\rangle$ into a DF-$\mathcal{ALC}$ ontology $\Gamma=\langle\mathcal{T}',\mathcal{A}\rangle$. Firstly, normalize ontology $\mathcal{O}$ into $\mathcal{O}'$ where only TBox of $\mathcal{O}$ and the concept name set changed. Assign $\mathcal{T}'$ to the TBox of $\Gamma$. Fuzzy extend $\mathcal{A}$ to be the ABox of of $\Gamma$ (i.e. transform each assertion $\phi$ in $\mathcal{A}$ to be $\phi=1$).
The signature of $\Gamma$ is defined as the signature of $\mathcal{O}'$, containing $N_C$, $N_R$, and $N_I$.
To enable differentiable operators to transfer gradient information, we reform the fuzzy interpretation as differentiable fuzzy interpretation. The domain of the grounding $\Delta^\mathcal{I}$ is $N_I$.
The interpretation function of $\mathcal{I}$ is reformed as an embedding function, which embeds each concept name $C\in N_C$ into $|\Delta|$-dimensional vector, $C^{\mathcal{I}}=\mathbb{R}^{|\Delta|}$, and each role name $r\in N_R$ into a $(|\Delta|, |\Delta|)$-dimensional matrix, $r^{\mathcal{I}}=\mathbb{R}^{|\Delta|}\times\mathbb{R}^{|\Delta|}$. The $i$th item of $C^{\mathcal{I}}$ is the truth value of $\Delta_i:C$, and the $(i,j)$th item of $r^{\mathcal{I}}$ is the truth value of $(\Delta_i, \Delta_j):r$.

\subsection{Semantics}
DF-$\mathcal{ALC}$ is a member of differentiable fuzzy logics~\citep{van2022analyzing}, which means the truth values of axioms are continuous, logical operators are interpreted into differentiable fuzzy operators, and the interpretation function is an embedding function. Except for the differentiable property, in semantics, DF-$\mathcal{ALC}$ is the same with fuzzy $\mathcal{ALC}$. 
Given DF-$\mathcal{ALC}$ ontology $\Gamma$, the properties that DF-$\mathcal{ALC}$ holds are shown in Table~\ref{tab:properties}. 
Only $C\sqcap\neg C\cong\bot$ and $C\sqcup\neg C\cong\top$ do not hold in DF-$\mathcal{ALC}$ and fuzzy $\mathcal{ALC}$.
But for any $C$ and $\mathcal{I}$, $(C\sqcup\neg C)^{\mathcal{I}}\geq 0.5$ and $(C\sqcap\neg C)^{\mathcal{I}}\leq 0.5$ hold.
\begin{table}[ht]
    \centering
    \resizebox{0.4\textwidth}{!}{
    \begin{tabular}{@{}ccc@{}}
    \toprule
      Property & Fuzzy $\mathcal{ALC}$ & Real Logic \\ \midrule
      
       $C\sqcap\neg C\cong\bot$ &  & \\
       $C\sqcup\neg C\cong\top$ &  & \\
       $C\sqcap C\cong C$ & $\bullet$& \\
       $C\sqcup C\cong C$ & $\bullet$& \\
       $\neg\neg C\cong C$ & $\bullet$& $\bullet$\\
       $\neg (C\sqcap D)\cong\neg C\sqcup\neg D$ & $\bullet$& \\
       $\neg (C\sqcup D)\cong\neg C\sqcap\neg D$ & $\bullet$& \\
       $C\sqcap(D\sqcup E)\cong (C\sqcap D)\sqcup (C\sqcap E)$ & $\bullet$& \\
       $C\sqcup(D\sqcap E)\cong (C\sqcup D)\sqcap (C\sqcup E)$ & $\bullet$& \\
       $\forall r.C\cong\neg\exists r.\neg C$ & $\bullet$& \\
       
       \bottomrule
       
    \end{tabular}
    }
    \caption{Properties of fuzzy $\mathcal{ALC}$ with equality assertion used in our model and product real logic~\citep{van2022analyzing} used in LTN~\citep{badreddine2022logic}. Cell with $\bullet$ is the meaning of `has'.}
    \label{tab:properties}
\end{table}

The semantics of DF-$\mathcal{ALC}$ is \emph{sound} w.r.t crisp semantics under the open-world assumption. This is an extension of the soundness of fuzzy $\mathcal{ALC}$. We define the crisp transformation $\sharp(\cdot)$ of DF-$\mathcal{ALC}$ assertion $\varphi$ into three-valued $\mathcal{ALC}$ assertion.
\begin{equation}
    \sharp\varphi=\sharp\{\phi=n\} \mapsto \left\{\begin{array}{cc}
       \phi  &  \mathrm{when\;} n>\alpha\\
        unknown &  \mathrm{when\;} 1-\alpha\leq n\leq\alpha\\
        \neg\phi &  \mathrm{when\;} n< 1-\alpha\\
    \end{array}\right\}
    \label{eq:crispy}
\end{equation}
where $\alpha\in[0.5,1]$ is predefined according to the application, and $\neg\phi$ is $a:\neg C$ or $(a,b):\neg r$.
For TBox axioms, $\sharp(\cdot)$ to fuzzy TBox axioms: $\sharp\{\psi\in\mathcal{T}\}=\{\psi\in\mathcal{T}\}$. So for $\mathcal{O}=\langle\mathcal{T},\mathcal{A}\rangle$, $\sharp\mathcal{O}=\sharp\{\varphi\in\mathcal{A}\}\cup\{\psi\in\mathcal{T}\}$.

\begin{prop}(Soundness of the semantics)
\label{prop:1}
Let $\sharp\mathcal{O}$ be an $\mathcal{ALC}$ ontology, and $\varphi$ be a fuzzy assertion. $\mathcal{O}\vapprox\varphi$ iff. $\sharp\mathcal{O}\models\sharp\varphi$ (i.e. fuzzy entailment is consistent with entailment in $\mathcal{ALC}$).
\end{prop}
Proof. 1.$\Rightarrow$ Consider any crisp interpretation $\mathcal{I}$ that is a model of $\sharp\mathcal{O}$. $\mathcal{I}$ can also be considered as a fuzzy interpretation that $C^{\mathcal{I}}(a)\in\{0,0.5,1\}$ and $r^{\mathcal{I}}(a,b)\in\{0,0.5,1\}$ hold. By induction on the structure of a concept $C$, $\mathcal{I}$ satisfies $a:C$ iff $C^{\mathcal{I}}(a)=1$. $\mathcal{I}$ satisfies $a:C$ is unknown iff $C^{\mathcal{I}}(a)=0.5$. 
And similarly for roles. Therefore, $\mathcal{I}$ is also a model of $\mathcal{O}$. And for every model of $\sharp\mathcal{O}$, $\mathcal{I}$ satisfies $\sharp\varphi$, so $\sharp\mathcal{O}\models\sharp\varphi$ holds. 

2. $\Leftarrow$ If $\sharp\mathcal{O}\models\sharp\varphi$, consider the crisp interpretation $\mathcal{I}$ discussed above, it is similar to proof that each model $\mathcal{I}$ of $\sharp\mathcal{O}$, is also the model of $\mathcal{O}$, and satisfies $\varphi$, so we have $\mathcal{O}\vapprox\varphi$.
To sum up, this proposition is proven to be true.


\subsection{Learning to Ground Symbols}
Given an $\mathcal{ALC}$ ontology $\mathcal{O}$, we have introduced how to transform it into the DF-$\mathcal{ALC}$ ontology $\Gamma=\langle\mathcal{T}',\mathcal{A}\rangle$. Then, transform the perceptual grounding $\mathcal{I}'$ into a differentiable fuzzy interpretation of $\Gamma$, as the initialization of the grounding of $\Gamma$.
Backpropagation on the grounding of $\Gamma$ can learn a model $\mathcal{I}''$ of $\Gamma$, which is also a model of $\mathcal{O}$ in the signature of $\mathcal{O}$, according to Theorem~\ref{theorem:normalization}. The forward process is to compute the truth values of axioms in $\mathcal{T}'$, by maximizing the satisfiability of $\mathcal{T}'$ (minimizing the \emph{hierarchical loss} in Equation ~\ref{e:7}).

\begin{equation}
    Loss(\mathcal{I}, \Gamma) = \frac{1}{|\mathcal{T}'|}\sum_{\{C\sqsubseteq D\}\in\mathcal{T}'} \sum_{a\in\Delta^{\mathcal{I}}}\mathrm{max}(0,C^{\mathcal{I}}(a)-D^{\mathcal{I}}(a))
    \label{e:7}
\end{equation}where $C$ and $D$ is any concept.

The main idea of Equation ~\ref{e:7} is to ensure that the interpretation $\mathcal{I}$ should satisfy every $C\sqsubseteq D\in\mathcal{T}'$, denoting that $\forall a\in \Delta^{\mathcal{I}}, C^{\mathcal{I}}(a)\leq D^{\mathcal{I}}(a)$.
Though $C\sqsubseteq D$ is also equivalent to $C\rightarrow D\geq n$ in fuzzy $\mathcal{ALC}$, where $(C\rightarrow D)^{\mathcal{I}} = \mathrm{min}_{d\in\Delta^{\mathcal{I}}}\{\mathrm{max}\{1-C^{\mathcal{I}}(a),D^{\mathcal{I}}(a)\}\}$, according to \citep{straccia2001reasoning}, it is hard to assign $n$. Besides, using $C\rightarrow D\geq n$ as a constraint will lead to $D^{\mathcal{I}}(a)\geq n$ or $C^{\mathcal{I}}(a)\leq 1-n$, which is highly dependent on $n$ rather than the reliable observations of the neural system. Besides, it is hard to decide whether to compel $D^{\mathcal{I}}(a)\geq n$ or $C^{\mathcal{I}}(a)\leq 1-n$. So we use $\forall a\in \Delta^{\mathcal{I}}, C^{\mathcal{I}}(a)\leq D^{\mathcal{I}}(a)$ to constraint concept inclusion in $\Gamma$.

\begin{prop}
\label{prop:2}
(\textbf{Soundness of learning to ground in DF-$\mathcal{ALC}$})  When the hierarchical loss converges to 0, the learned interpretation $\mathcal{I}''$ is the model of the given $\mathcal{ALC}$ ontology $\mathcal{O}$. For any model $\mathcal{J}$ of $\mathcal{O}$, $\mathcal{I}''\sim_{sig(\mathcal{O})}\mathcal{J}$.
\end{prop}
Proof. when loss converges to 0, the learned $\mathcal{I}''$ satisfies any $C\sqsubseteq D$ in the normalized ontology $\mathcal{O}'$, so $\mathcal{I}''$ is the model of $\mathcal{O}'$. And according to Theorem~\ref{theorem:normalization}, any model of $\mathcal{O}'$  is semantically equivalent to the model of $\mathcal{O}$. So this proposition is proved to be true.

\begin{table*}[ht]
    \centering
    \resizebox{0.9\textwidth}{!}{
    \begin{tabular}{@{}ccccc@{}}
    \toprule
      Example & Description &  Axiom & Expected & Performance of Hierarchical Loss \\ \midrule
      
      \ref{ex:1} &  $(\texttt{r}^{\mathcal{I}}(s_1,s_2)>\alpha)\land (\texttt{B}^{\mathcal{I}}(s_1)>\alpha)\implies (\texttt{A}^{\mathcal{I}}(s_2)>\alpha)$  & $\forall\texttt{r}.\texttt{A}\sqsubseteq \texttt{B}$, $\exists\texttt{r}.\texttt{A}\sqsubseteq \texttt{B}$  & Unknown & do not revise, as expected \\
      \ref{ex:1}& $(\texttt{r}^{\mathcal{I}}(s_1,s_2)>\alpha)\land (\texttt{B}^{\mathcal{I}}(s_1)>\alpha)\implies (\texttt{A}^{\mathcal{I}}(s_2)>\alpha)$ & $\texttt{B}\sqsubseteq \forall\texttt{r}.\texttt{A}$, $ \texttt{B}\sqsubseteq\exists\texttt{r}.\texttt{A}$ & True & can revise, but not in an expected way \\
      \ref{ex:2}& $(\texttt{r}^{\mathcal{I}}(s_1,s_2)>\alpha)\land (\texttt{A}^{\mathcal{I}}(s_1)>\alpha)\implies (\texttt{B}^{\mathcal{I}}(s_2)>\alpha)$ & $\forall\texttt{r}.\texttt{A}\sqsubseteq \texttt{B}$, $\exists\texttt{r}.\texttt{A}\sqsubseteq \texttt{B}$  & True & can revise,  but not in an expected way \\
      \ref{ex:2}& $(\texttt{r}^{\mathcal{I}}(s_1,s_2)>\alpha)\land (\texttt{A}^{\mathcal{I}}(s_1)>\alpha)\implies (\texttt{B}^{\mathcal{I}}(s_2)>\alpha)$ &  $\texttt{B}\sqsubseteq \forall\texttt{r}.\texttt{A}$, $ \texttt{B}\sqsubseteq\exists\texttt{r}.\texttt{A}$ & Unknown & can revise, but not in an expected way\\
       \ref{ex:3}&$(\texttt{A}^{\mathcal{I}}(s_2)>\alpha)\land (\texttt{B}^{\mathcal{I}}(s_1)>\alpha)\implies (\texttt{r}^{\mathcal{I}}(s_1,s_2)>\alpha)$ &  $\texttt{B}\sqsubseteq\exists\texttt{r}.\texttt{A}$ & True & \makecell{do not revise, as expected} \\
       \ref{ex:3}&$(\texttt{A}^{\mathcal{I}}(s_2)>\alpha)\land (\texttt{B}^{\mathcal{I}}(s_1)>\alpha)\implies (\texttt{r}^{\mathcal{I}}(s_1,s_2)>\alpha)$ & $\exists\texttt{r}.\texttt{A}\sqsubseteq \texttt{B}$, $\forall\texttt{r}.\texttt{A}\sqsubseteq \texttt{B}$ & Unknown & \makecell{do not revise, not as expected} \\
       \bottomrule
       
    \end{tabular}
    }
    \caption{Performance of hierarchical loss based on four kinds of axioms given the perceptual grounding in three examples. Unknown situations expect no revision. True situations expect revision executed in the implication. }
    \label{tab:examples}
\end{table*}

\subsection{Rule-based Learning}
\label{sec:rl}
But learning grounding by maximizing the satisfiability captured by DF-$\mathcal{ALC}$ semantics cannot effectively reason to revive the perceptual grounding. Standard reasoning in fuzzy description logic is to decide whether $\mathcal{O}\vapprox \varphi$ holds and  in crisp description logic to decide whether $\mathcal{O}\models\phi$ holds, which means to check whether all the models of $\mathcal{O}$ satisfy $\varphi$ or $\phi$. So the grounding considered in the standard reasoning is the models of knowledge base $\mathcal{O}$. While reasoning to ground is to revise perceptual grounding as a model of knowledge base $\mathcal{O}$, where the grounding contains plausible and implausible parts. The hierarchical loss proposed in Equation.~\ref{e:7} can lead $\texttt{A}\sqsubseteq\texttt{B}$ to learn a grounding $\mathcal{I}$ that $\texttt{A}^{\mathcal{I}}=\texttt{B}^{\mathcal{I}}$, so cannot be distinguish between $\sqsubseteq$ and $\equiv$. Besides, $\texttt{B}^{\mathcal{I}}(s)>\alpha$ can be revised as $0.5$, which losses information.

A relaxed revision is to reduce $\texttt{A}^{\mathcal{I}}(s)$ or improve less than $\texttt{B}^{\mathcal{I}}(s)$ to satisfy $\texttt{A}^{\mathcal{I}}\leq \texttt{A}^{\mathcal{I}}$. 
Here is the rule-based loss for axioms in the normal form 1-3 shown in Figure.~\ref{fig:normalized-forms}:

\begin{equation}
\begin{split}
    Loss_{\texttt{NF1-NF3}}(\texttt{A}^{\mathcal{I}},\texttt{B}^{\mathcal{I}};\mathcal{I}, \Gamma) = &\sum_{\texttt{A}\sqsubseteq \texttt{B}}\sum_{s\in \Delta^\mathcal{I}}((1-\texttt{B}^{\mathcal{I}}(s))*\\ & G(\texttt{A}^{\mathcal{I}}(s),\texttt{B}^{\mathcal{I}}(s)) 
\end{split}
\label{eq:8}
\end{equation}, where $\texttt{A}$ and $\texttt{B}$ is any concept, $\texttt{A}^{\mathcal{I}}$ and $\texttt{B}^{\mathcal{I}}$ are in the representation of DF-$\mathcal{ALC}$ calculated according to the semantics of fuzzy $\mathcal{ALC}$ based on G\"{o}del logic. $G(v,w) = ReLU(v-w)$, and $G(v,w)$ does not take part in the gradient descent. 

For axioms in the normal form 4-7, the semantics of $\exists$ and $\forall$ in fuzzy $\mathcal{ALC}$ based on G\"{o}del logic can not be reasoned in a proper way with hierarchical loss. So different losses are designed for axioms in normal form 4-7 respectively.

Consider four ontologies $\mathcal{O}_1 = \{\exists\texttt{r}.\texttt{A}\sqsubseteq \texttt{B}\}$, $\mathcal{O}_2 = \{\forall\texttt{r}.\texttt{A}\sqsubseteq \texttt{B}\}$, $\mathcal{O}_3 = \{\texttt{B}\sqsubseteq\exists\texttt{r}.\texttt{A}\}$, $\mathcal{O}_4 = \{\texttt{B}\sqsubseteq\forall\texttt{r}.\texttt{A}\}$, in the  following three examples, the performance of DF-$\mathcal{ALC}$ with hierarchical loss and the revision calculus are shown in Table.~\ref{tab:examples}. 


\begin{example}
\label{ex:1}
Given a perceptual grounding $\mathcal{I}$ in the domain $\{s_1, s_2\}$, $\texttt{A}^{\mathcal{I}}(s_1)=0,\texttt{A}^{\mathcal{I}}(s_2)=0$, $\texttt{B}^{\mathcal{I}}(s_1)=0.9,\texttt{B}^{\mathcal{I}}(s_2)=0$, $\texttt{r}^{\mathcal{I}}(s_1,s_2)=0.9,\texttt{r}^{\mathcal{I}}(s_1,s_1)=\texttt{r}^{\mathcal{I}}(s_2,s_1)=\texttt{r}^{\mathcal{I}}(s_2,s_2)=0$, notated as vectors
 $\texttt{A}^{\mathcal{I}}=[0,0]$, $\texttt{B}^{\mathcal{I}}=[0.9,0]$, and matrix \begin{equation*}
\texttt{r}^{\mathcal{I}} = 
\begin{bmatrix}
0 & 0.9  \\
0 & 0  
\end{bmatrix}
\end{equation*}. 
\end{example} 
According to the semantics of fuzzy $\mathcal{ALC}$, 
in $\mathcal{O}_1$, $(\exists  \texttt{r}.\texttt{A})^{\mathcal{I}}=[0,0]$, which satisfies $(\exists \texttt{r}.\texttt{A})^{\mathcal{I}}\leq \texttt{B}^{\mathcal{I}}$, so hierarchical loss is 0, and no revision is executed. But this is not what we want. As we know that $s_1$ is likely to be $\texttt{B}$, and $\texttt{r}(s1,s2)$ is likely to be true, so $s_2$ is likely to be a membership of $\texttt{A}$.
In $\mathcal{O}_2$, $(\forall  \texttt{r}.\texttt{A})^{\mathcal{I}}=[0.1,1]$, which does not satisfy $(\forall \texttt{r}.\texttt{A})^{\mathcal{I}}\leq \texttt{B}^{\mathcal{I}}$, and hierarchical loss is $1.1$. Through gradient decent, until loss becomes $0$, $\texttt{A}^{\mathcal{I}}=[0.24,0]$,$\texttt{B}^{\mathcal{I}}=[0.4,1]$, $\texttt{r}^{\mathcal{I}}(s_2,s_1)$ will be $0.7$ and $\texttt{r}^{\mathcal{I}}(s_1,s_2)$ will be $1$.
In $\mathcal{O}_3$, with hierarchical loss, $\mathcal{A}^{\mathcal{I}}$ will be $[0.38,0]$ , $\mathcal{B}^{\mathcal{I}}$ will be $[0.35,0]$and $\texttt{r}^{\mathcal{I}}(s_1,s_1)$ will be $[0,36]$.
In $\mathcal{O}_4$, with hierarchical loss, $\mathcal{B}^\mathcal{I}=[0,0]$ and $\texttt{r}^{\mathcal{I}}(s_1,s_2)=0$ 
$\mathcal{O}_4$
\begin{example}
\label{ex:2}
Given a perceptual grounding $\mathcal{I}$ in the domain $\{s_1, s_2\}$, $\texttt{A}^{\mathcal{I}}=[0,0.9]$, $\texttt{B}^{\mathcal{I}}=[0,0]$, $\texttt{r}^{\mathcal{I}}$ is the same as in Example. 1.
\end{example}
 According to the semantics of fuzzy $\mathcal{ALC}$, in $\mathcal{O}_1$, $(\exists  \texttt{r}.\texttt{A})^{\mathcal{I}}=[0.9,0]$, which does not satisfy $(\exists \texttt{r}.\texttt{A})^{\mathcal{I}}\leq \texttt{B}^{\mathcal{I}}$, so hierarchical loss is 0.9. Through gradient decent, until loss becomes 0, $\texttt{A}^{\mathcal{I}}$ is decreased as $[0,0]$, and $\texttt{B}^{\mathcal{I}}$ is increased as $[0.9,0]$. $\texttt{A}^{\mathcal{I}}$ is not expected to be changed and $\texttt{B}^{\mathcal{I}}$ is expected to be increased as $[0.9,0]$.
 In $\mathcal{O}_2$, $(\forall  \texttt{r}.\texttt{A})^{\mathcal{I}}=[0.9,1]$, $\texttt{A}^{\mathcal{I}}$ will be revised as $[0,0]$, $\texttt{B}^{\mathcal{I}}$ will be revised as $[0.5,1]$, and $\texttt{r}^{\mathcal{I}}(s_1,s_2)=1$.
 In $\mathcal{O}_3$ and $\mathcal{O}_4$, there is no revision.


\begin{example}
\label{ex:3}
Given a perceptual grounding $\mathcal{I}$ in the domain $\{s_1, s_2\}$, $\texttt{A}^{\mathcal{I}}=[0,0.9]$ and $\texttt{B}^{\mathcal{I}}=[0.9,0]$, \begin{equation*}
\texttt{r}^{\mathcal{I}} = 
\begin{bmatrix}
0 & 0  \\
0 & 0  
\end{bmatrix}.
\end{equation*} 
\end{example}
According to the semantics of fuzzy $\mathcal{ALC}$, $(\exists  \texttt{r}.\texttt{A})^{\mathcal{I}}= (\forall \texttt{r}.\texttt{A})^{\mathcal{I}}=[0,0]$, which satisfies $(\exists \texttt{r}.\texttt{A})^{\mathcal{I}}\leq \texttt{B}^{\mathcal{I}}$, so no revision is executed. 
In $\mathcal{O}_3$, $\texttt{B}^{\mathcal{I}}$ and $\texttt{r}^{\mathcal{I}}$ is revised if there are other $s_n$ that $\texttt{A}^{\mathcal{I}}$ is not zero.
In  $\mathcal{O}_4$, there is no revision.

 
We introduce the following rule-based loss to solve the problem meets in Table.~\ref{tab:examples}:
\begin{equation}
\begin{split}
    &Loss_{\texttt{NF4}}(\texttt{A}^{\mathcal{I}},\texttt{r}^{\mathcal{I}};\mathcal{I}, \Gamma) = \sum_{\texttt{B}\sqsubseteq\exists\texttt{r}.\texttt{A}}\sum_{s\in \Delta^\mathcal{I}}((1-\texttt{A}^{\mathcal{I}}(s))*G(\alpha',\texttt{A}^{\mathcal{I}}(s))*\\  &G(\sum_{a\in\Delta^{\mathcal{I}}}(\texttt{B}^{\mathcal{I}}(a)\otimes\texttt{r}^{\mathcal{I}}(a,s),\alpha'))+
    (1-\texttt{r}^{\mathcal{I}}(s,a))*\\&
    G(\texttt{B}^{\mathcal{I}}(s)\otimes\texttt{A}^{\mathcal{I}}(a),\texttt{r}^{\mathcal{I}}(s,a)))
\end{split}
\label{eq:11}
\end{equation} where $\otimes$ is the t-norm. With the rule-based loss, the parts of interpretation that we believed to be true are not determined by a threshold, but by the specialty of the task and dataset. In this paper, we use the product t-norm as $\otimes$ in the rule-based loss validated by the evaluation in the experiments. $\alpha'\in[0.5,1]$ is the threshold for the truth value.
\begin{equation}
\begin{split}
    Loss_{\texttt{NF5}}(\texttt{A}^{\mathcal{I}},\texttt{r}^{\mathcal{I}};\mathcal{I}, \Gamma) = &\sum_{\texttt{B}\sqsubseteq\forall\texttt{r}.\texttt{A}}\sum_{s\in \Delta^\mathcal{I}}((1-\texttt{A}^{\mathcal{I}}(s))*G(\alpha,\texttt{A}^{\mathcal{I}}(s))*\\ & G(\sum_{a\in\Delta^{\mathcal{I}}}(\texttt{B}^{\mathcal{I}}(a)\otimes\texttt{r}^{\mathcal{I}}(a,s),\alpha)))
\end{split}
\label{eq:12}
\end{equation}
\begin{equation}
\begin{split}
    Loss_{\texttt{NF6}}(\texttt{B}^{\mathcal{I}};\mathcal{I}, \Gamma) = &\sum_{\exists\texttt{r}.\texttt{A}\sqsubseteq \texttt{B}}\sum_{s\in \Delta^\mathcal{I}}((1-\texttt{B}^{\mathcal{I}}(s))*G(\alpha',\texttt{B}^{\mathcal{I}}(s))*\\ & G(\sum_{a\in\Delta^{\mathcal{I}}}(\texttt{A}^{\mathcal{I}}(a)\otimes\texttt{r}^{\mathcal{I}}(s,a),\alpha'))) \\
\end{split}
\label{eq:13}
\end{equation}
\begin{equation}
\begin{split}
    Loss_{\texttt{NF7}}(\texttt{A}^{\mathcal{I}};\mathcal{I}, \Gamma) = &\sum_{\forall\texttt{r}.\texttt{A}\sqsubseteq \texttt{B}}\sum_{s\in \Delta^\mathcal{I}}((1-\texttt{B}^{\mathcal{I}}(s))*G(\alpha',\texttt{B}^{\mathcal{I}}(s))*\\ & G(\sum_{a\in\Delta^{\mathcal{I}}}(\texttt{A}^{\mathcal{I}}(a)\otimes\texttt{r}^{\mathcal{I}}(s,a),\alpha'))+\\&
    (1-\texttt{A}^{\mathcal{I}}(s))*G(\alpha',\texttt{A}^{\mathcal{I}}(s))*\\ & G(\sum_{a\in\Delta^{\mathcal{I}}}(\texttt{B}^{\mathcal{I}}(a)\otimes\texttt{r}^{\mathcal{I}}(a,s),\alpha'))) \\
\end{split}
\label{eq:14}
\end{equation}

We only consider the situations when an assertion is larger than $\alpha'$ here,w.l.o.g., the opposite situations (less than $1-\alpha'$) are dual and can be added to the loss according to the distribution of perceptual grounding, e.g. the opposite situations are more plausible.


\begin{table*}[ht]
\centering
\resizebox{\textwidth}{!}{
    \begin{tabular}{@{}ccccccccccccccccc@{}}
    \toprule
    
    \multicolumn{1}{c}{} & \multicolumn{4}{c}{0.2} & \multicolumn{4}{c}{0.4}& \multicolumn{4}{c}{0.6} & \multicolumn{4}{c}{0.8} \\ \cmidrule(r){2-5}\cmidrule(r){6-9}\cmidrule(r){10-13}\cmidrule(r){14-17}
    \multicolumn{1}{c}{} & M & D w/HL & D w/RL  & \multicolumn{1}{c}{L} & M & D w/HL & D w/RL& \multicolumn{1}{c}{L} & M & D w/HL & D w/RL & \multicolumn{1}{c}{L} & M & D w/HL & D w/RL & L \\ \midrule
Family & 0.0 & \textbf{100.0} & 93.1 & \textbf{100.0} & 0.0 & \textbf{100.0} & 93.1 & \textbf{100.0} & 0.0 & \textbf{100.0} & 93.1& \textbf{100.0} & 0.0 & \textbf{100.0} & 93.1& \textbf{100.0} \\
Family2 & 0.0 & \textbf{100.0}& 73.8 & 91.8 & 0.0 & \textbf{96.7}& 73.8 & 88.5 & 0.0 & \textbf{98.4}& 73.8 & 91.8 & 0.0 & \textbf{96.7}& 73.8 & 91.8 \\
GlycoRDF & 4.1 & \textbf{100.0}& 90.9 & 96.4 & 4.1 & \textbf{100.0}& 90.9 & 96.4 & 4.1 & \textbf{99.5}& 90.9 & 95.9 & 4.1 & \textbf{99.5}& 90.9 & 95.9 \\
Nifdys & 7.3 & \textbf{97.3}& 89.2 & 95.0 & 2.7 & \textbf{98.7}& 84.5 & 94.7 & 0.4 & \textbf{99.3}& 82.7 & 94.8 & 0.1 & \textbf{99.6}& 82.7 & 94.7 \\
Nihss & 16.1 & \textbf{100.0}& \textbf{100.0} & 51.6 & 0.0 & \textbf{100.0}& \textbf{100.0 }& 48.4 & 0.0 & \textbf{100.0}& \textbf{100.0} & 48.4 & 0.0 & \textbf{100.0}& \textbf{100.0} & 48.4 \\
Ontodm & 5.4 & 91.6& 41.5 & \textbf{98.5} & 0.3 & 91.2& 37.3 & \textbf{98.2} & 0.2 & 92.1& 37.3 & \textbf{97.6} & 0.2 & 91.3& 37.3 & \textbf{97.4} \\
Sso & 0.0 & \textbf{100.0}& \textbf{100.0} & \textbf{100.0} & 0.0 & \textbf{100.0}&  \textbf{100.0} & \textbf{100.0} & 0.0 & \textbf{100.0}& \textbf{100.0} & \textbf{100.0} & 0.0 & \textbf{100.0}& \textbf{100.0} & \textbf{100.0} \\  \bottomrule
    \end{tabular}
}
\caption{Success rate (\%) in four mask rate settings with $\alpha=0.8$. M is the masked grounding. D is the DF-$\mathcal{ALC}$ revised grounding based on M (w/HL is with hierarchical loss, w/RL is with rule-based loss). L is the LTN revised grounding based on M.}
\label{tab:result1}
\end{table*}

\begin{figure*}[ht]
    \centering
     \includegraphics[width=0.8\linewidth]{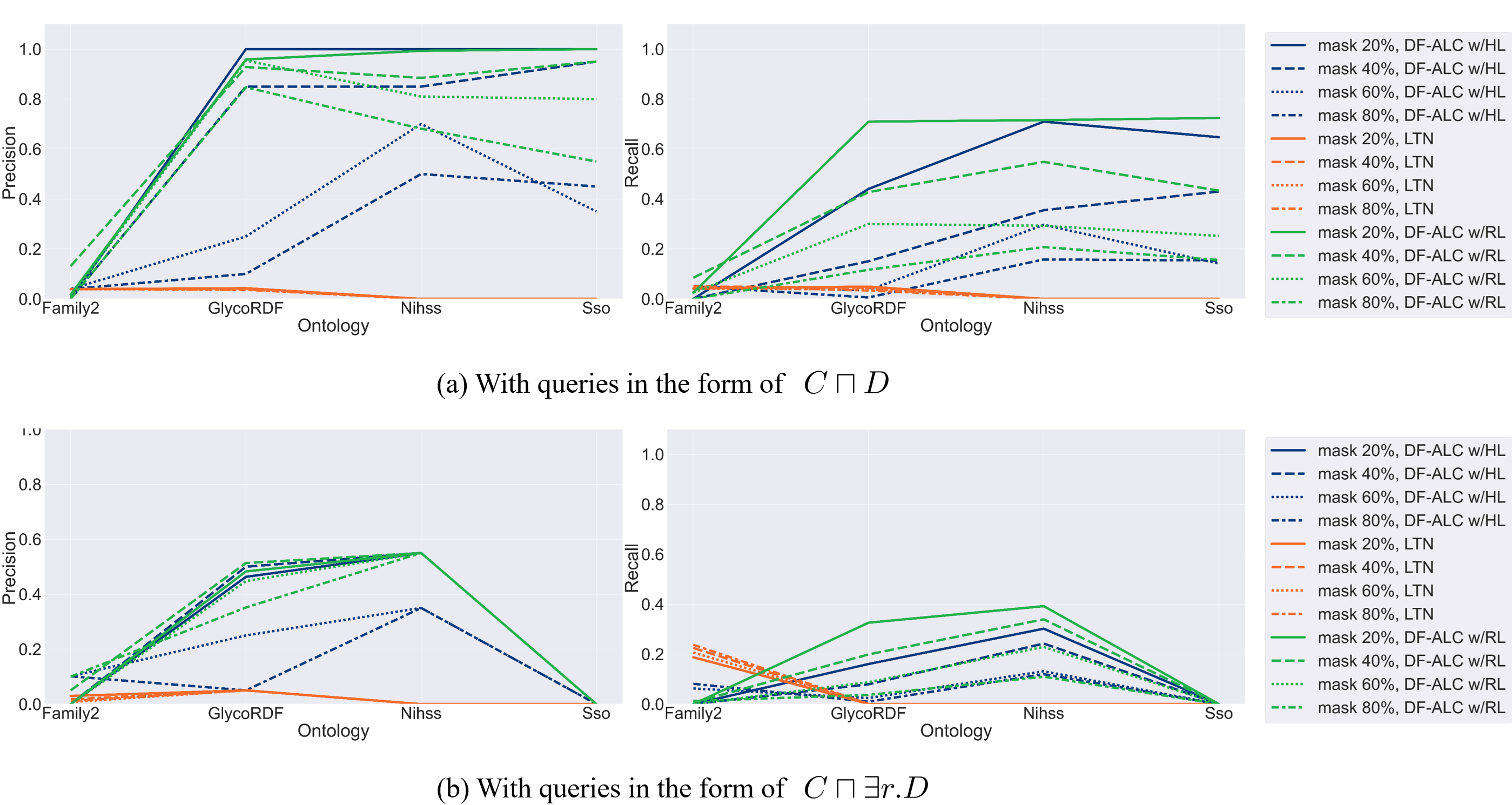}
     \caption{Conjunctive query answering results.}
     \label{fig:cqa_result}
\end{figure*}

\begin{table}
\centering
\resizebox{0.5\textwidth}{!}{
\begin{tabular}{@{}cccccccc@{}}
\toprule
   & Family & Family2 &  GlycoRDF  &  Nifdys& Nihss& Ontodm& Sso\\\midrule
   \# TBox axioms & 2032 & 2054& 1453& 6435& 318& 3476& 2050\\
   \# ABox axioms  & 224 & 224& 518& 2920& 146& 1113& 366\\
   \# Concepts & 19& 19& 113& 2751& 18& 838& 176\\
   \# Roles & 4& 4& 91& 68& 16& 78& 22\\
   \# Individuals & 202& 202& 219& 102& 106& 187& 158 \\
   Expressivity & $\neg$ & $\neg$, $\sqcap$, $\exists$& $\neg$, $\sqcup$, $\exists$ & $\neg$, $\sqcup$, $\exists$ & $\neg$&$\neg$, $\sqcap$, $\sqcup$, $\exists$, $\forall$ & /\\
   \bottomrule

\end{tabular} }
\caption{Ontology information}
 \label{tab:info}
\end{table}
\section{Experiments}
\subsection{Performance Evaluation}
We design two experiments to verify the efficiency of DF-$\mathcal{ALC}$, and answer the following research questions (RQ): 
\begin{itemize}
    \item RQ1: Can hierarchical loss always converge to zero? If not, what can the result be in these cases?
    \item RQ2: How successful the learning in DF-$\mathcal{ALC}$ is in keeping reliable observations while revising erroneous ones?
    \item RQ3: How does rule-based loss perform compared to hierarchical loss?
\end{itemize}

To answer these questions, learning grounding for DF-$\mathcal{ALC}$ ontologies should be evaluated in different neural networks under various situations. But the distribution of observations and the properties of a specific neural network gives bias (in a way of having the same pattern of errors) to the perceptual grounding. Therefore, we design an experimental task --- masked ABox revision, for evaluation in various observation distributions. This task is not oriented to tackle a concrete symbol grounding problem. Given an $\mathcal{ALC}$ ontology $\mathcal{O}=\langle\mathcal{T},\mathcal{A}\rangle$, where $\mathcal{A}$ is completed by a logical reasoner, then fuzzy extended in DF-$\mathcal{ALC}$. We assign the ideal grounding $\mathcal{I}$ with the processed $\mathcal{A}$. Mask the random part of grounding $\mathcal{I}$ into a random truth value in an unknown region, and reformulate it into a differentiable fuzzy interpretation $\mathcal{I}'$ as the imitation of a perceptual grounding. Then transform ontology $\mathcal{O}$ into DF-$\mathcal{ALC}$ ontology $\Gamma$. Use $\mathcal{I}'$ as the initialization to learn the revised grounding $\mathcal{I}''$ based on $\Gamma$.
Using the crisp transformation defined in Formula~\ref{eq:crispy} with $\alpha=0.5$ to transform the revised grounding into crisp grounding, evaluate it with the satisfiability calculated in the crisp mode.  However, solely evaluating the satisfiability of $\mathcal{O}$ with $\mathcal{I}''$ cannot show the constancy (in keeping the reliable parts) between $\mathcal{I}'$ and $\mathcal{I}''$, as one ontology is satisfiable in multiple groundings.
So we design another task called conjunctive query answering, which is a kind of ontology-mediated query answering. The target is to retrieve individuals for complex concepts based on the revised grounding.


\subsubsection{Settings}
In the masked ABox revision task, we used 6 ontologies (``Ontodm'' and ``Nifdys'' are not consistent), while in the conjunctive query answering task, we used 4 consistent ontologies.

The mask rate of ABox ranges from \{20\%, 40\%, 60\%, 80\%\}. We set the unknown region as $[0.2,0.8]$. 
Meanwhile, the truth values greater (less) than $\alpha=0.8$ ($1-\alpha=0.2$) were assumed to be true (false). 
We used the Logical Tensor Network (LTN) as the comparison model. LTN is a differentiable fuzzy logic model in product real logic based on first-order logic. But LTN needs to pre-train predicates and functions with labeled data. To adapt LTN to this task, we removed the parameters of predicates and functions and used the hierarchical loss to train LTN.
Besides, we trained DF-$\mathcal{ALC}$ with rule-based loss ($\alpha'=0.8$). These two compared models share the same masked grounding in different settings. 

We used success rate (S.R.) to evaluate soundness. The success rate is the percentage of the TBox axiom in original ontology (without normalization) that is satisfied w.r.t. the crisp $\mathcal{I}''$. To be fair, we evaluated the results in the semantics of first-order logic. We tested conjunctive queries in the forms of $C\sqcap D$ and $C\sqcap \exists r.D$, where $C$ and $D$ are atomic concepts, and $r$ is a role name. We generated 20 queries in each form, and the answer set of each query was not empty. Considering the time complexity of using a logical reasoner to get the true answer set, we only used two forms of conjunctive queries (CQs) in-depth 2 (the depth is determined by the conjunction amounts in the query). We chose all individuals with $Q^{\mathcal{I}}(a)\geq 0.8$ to be the answer for query $Q$. And use the answers generated by logical reasoner as ideal answers to evaluate the predicted answers with precision and recall as metrics. 

The ontologies used for the experiments are taken from Bioportal\footnote{http://bioportal.bioontology.org/ontologies}, which, currently, includes more than 700 biomedical ontologies from different sources. We require the ontologies to have at least the logical operator of negation, disjunction, or universal quantifier, as well as 100 ABox assertions. Five ontologies fall into this set, with two of them (``Ontodm'' and ``Nifdys'') not consistent in some assertions; it remains to see whether DF-$\mathcal{ALC}$ would revise these errors. A taxonomy ontology (Sso) is also added for comparison. We also test a terseness ontology ``Family'', which contains multiple instantiated families but its knowledge is incomplete. Based on ``Family'', we augment it into ``Family2'' by adding some knowledge that can bridge with the instantiation. The information about these ontologies is shown in Table~\ref{tab:info}
Adam optimizer was used with a learning rate of 2e-4 to learn the grounding. Early stopping with 10 epochs tolerance was used to limit the running time.

\begin{figure*}[ht]
    \centering
     \includegraphics[width=\linewidth]{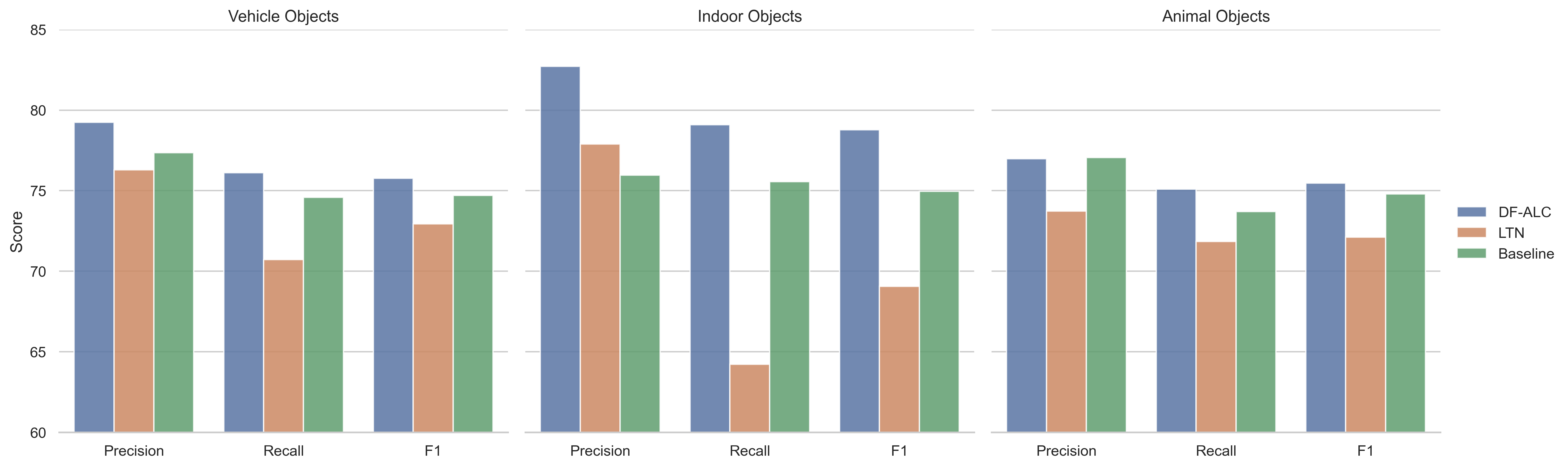}
     \caption{Object types classification results. The baseline model is the classification results of FRCNN. For evaluation, the performance is classified into three categories: vehicle, indoor, and animal. The metrics are macro averaged.}
     \label{fig:sii_result}
\end{figure*}

\subsubsection{Results}
\label{sec:experiment-results}

From the results shown in Table ~\ref{tab:result1}, where we can see that DF-$\mathcal{ALC}$ (w/HL) and LTN succeeded in most cases.
Not surprisingly, the success rate is low for masked grounding, since any small fault in the grounding can dissatisfy an axiom in the ontology. 
DF-$\mathcal{ALC}$ (w/HL) does not perfectly ground ``Ontodm'' and ``Nifdys'', as these two ontologies are not consistent. In ``Ontodm'', DF-$\mathcal{ALC}$ (w/HL) predicts wrongly disjoint concepts, and these concepts are incompletely asserted. The same problem occurs in ``Nifdys''. We further study the failures in ``Family2'', and find that the failures are caused by unknown cases. More specifically, we can see that an individual ``F6M80'' is asserted as a Male, but his parents are not asserted, therefore the values of ``Son(F6M80)'' and ``Child(F6M80)'' are unknown. In learned grounding of $\Gamma$, though they are all in unknown region $(0.2,0.8)$, $\mathrm{Son}^{\mathcal{I}''}\mathrm{(F6M80)}=0.5490>\mathrm{Child}^{\mathcal{I}''}\mathrm{(F6M80)}=0.5489$ can still lead to $\mathrm{Son}^{\mathcal{I}''}\not\sqsubseteq\mathrm{Child}^{\mathcal{I}''}$. For LTN, almost all of the axioms in the form of $\exists r.\top\sqsubseteq C$ is not learned well. For DF-$\mathcal{ALC}$, several axioms fail in the complex forms ( e.g. $\exists r_1.(\exists r_2.B) \sqsubseteq C$) when the masked rate gets higher. 
The success rate of DF-$\mathcal{ALC}$ (w/RL) is low in most cases, as in most of the ontologies, the rule-based loss cannot achieve zero. Besides, with rule-based loss, the semantics of quantifiers are not following fuzzy $\mathcal{ALC}$. 

To answer RQ1, it is worth noting that learning in ``Family2'', ``GlycoRDF'', ``Nifdys'', and ``Ontodm'' cannot get the hierarchical loss to converge to 0 in finite time in the four settings. But we still get the success rate of ``Family2'' and ``GlycoRDF'' being $100\%$, which is due to the crisp transformation for masked grounding. So if the given $\mathcal{ALC}$ ontology $\mathcal{O}$ is consistent, though the learning loss cannot converge to 0 in some cases, the crisp transformed grounding is the model of $\mathcal{O}$. 

 To answer RQ2, from the results shown in Figure ~\ref{fig:cqa_result}, we find that DF-$\mathcal{ALC}$ and LTN cannot do well in this task as expected. Because the masked semantics loses much information, DF-$\mathcal{ALC}$ can revise the grounding in a shifted direction. 
 High precision (i.e. remaining the reasoning properties well) and relatively high recall (i.e. being constancy with reliable part of observation) can be expected with DF-$\mathcal{ALC}$ compared to LTN when the mask rate is low (e.g. 0.2). While both DF-$\mathcal{ALC}$ and LTN have problems in revising role interpretation function. 

To answer RQ3, DF-$\mathcal{ALC}$ (w/RL) performs better than DF-$\mathcal{ALC}$ (w/HL) in the CQs masking task, while groundings trained by DF-$\mathcal{ALC}$ (w/RL) have less interpretability than those trained by DF-$\mathcal{ALC}$ (w/HL) when knowledge base contains axioms formed in NF 4-7. As rule-based loss deduces based on the assumption that grounding where the truth value is larger than $\alpha'$ is reliable. 

Overall, DF-$\mathcal{ALC}$ outperforms LTN in most observation revision cases. The common and significant problem for both of them is to avoid the disturbance of unknown cases to satisfiability to knowledge.

\subsection{Semantic Image Interpretation}
In this experiment, we apply DF-$\mathcal{ALC}$ (w/RL) to solve the SII problem, as DF-$\mathcal{ALC}$ (w/HL) cannot revise the grounding properly which has been explained in Section.~\ref{sec:rl} 

We use PASCAL-PART dataset~\cite{chen2014detect}. The dataset consists of images annotated with bounding boxes denoting distinct objects. The simple semantics between these objects like part-of relation can be detected by computing the pixel cover rate between bounding boxes, constructing the role interpretation function of perceptual interpretation $\mathcal{I'}$. Objects are then grounded by object detector Fast R-CNN (FRCNN)~\cite{girshick2015fast}, which gives each object $b$ the label $C$ with $score(C,b)$, constructing the concept interpretation function of $\mathcal{I'}$.
To revise $\mathcal{I'}$, we introduced an OWL ontology $\mathcal{O}_{partOf}$ with two kinds of axioms, which is similar to the ontology introduced in Figure.~\ref{fig:df-alc}. The first kind of axiom depicts the part-of relation between types, e.g. $\exists\texttt{isPartOf}.\texttt{Chair}\equiv \texttt{Seat}\sqcup\texttt{Leg}$. The second kind of axiom asserts the disjointness between different types, e.g. $\texttt{Chair}\sqcap\exists\texttt{isPartOf}.\texttt{Chair}\sqsubseteq\bot$, $\texttt{Chair}\sqcap\texttt{Table}\sqsubseteq\bot$. Then we used the rule-based loss to revise $\mathcal{I'}$ according to ontology $\mathcal{O}_{partOf}$. LTN was trained with constraints following the settings proposed in ~\cite{donadello2017logic}.

The results are shown in Figure.~\ref{fig:sii_result}. Indoor objects have the simplest relationships and animal objects have the most complex relationships, which interprets why DF-$\mathcal{ALC}$ performs the best in indoor objects. Animal objects can have many common types of objects, e.g. ear, head, and eye, but DF-$\mathcal{ALC}$ can still improve recall. LTN fails in improving the object types classification performance upon the baseline because the fuzzy logical operators cannot convey the proper information by maximizing the satisfiability. But LTN can do link prediction (e.g. revise the part-of-relation interpretation in this case), while current DF-$\mathcal{ALC}$ does not perform well in link prediction when the role grounding or concept grounding does terribly. Theoretically, DF-$\mathcal{ALC}$ (w/RL) can guide the revision of role interpretation with the knowledge formed as $\texttt{A}\sqsubseteq\exists r.\texttt{B}$ bridged by the similarities between individuals. On the whole, DF-$\mathcal{ALC}$ provides an unsupervised way to improve symbol grounding by utilizing logical knowledge compared to the existing methods which need supervised training before symbol grounding.

We also evaluated DF-$\mathcal{ALC}$ and LTN~\cite{donadello2017logic} in the low-resource SII task, shown in Table.~\ref{tab:lowresourse}. 
As LTN cannot promote performance, the results run with LTN are not shown here.
We can see that DF-$\mathcal{ALC}$ can promote the performance of baseline even when in low-resource cases.

\begin{table*}
\centering
\resizebox{\textwidth}{!}{
\begin{tabular}{@{}ccccccccccc@{}}
\toprule
\multicolumn{2}{c}{} & \multicolumn{3}{c}{Vehicle} & \multicolumn{3}{c}{Indoor}& \multicolumn{3}{c}{Animal}  \\ 
\cmidrule(r){3-5}\cmidrule(r){6-8}\cmidrule(r){9-11}
    Mask Rate& Model & P & R &  \multicolumn{1}{c}{F1}  & P & R &  \multicolumn{1}{c}{F1}  &P & R & F1 \\ \midrule
 0\% & baseline& 77.36 & 74.58 & 74.71 & 75.96 & 75.56 & 74.96 & 77.06 & 73.7 & 74.8\\
    0\% & DF-ALC (w/RL)& 78.35 ($\uparrow$ 0.99)& 75.83 ($\uparrow$ 1.25)& 75.97 ($\uparrow$ 1.26)& 80.69 ($\uparrow$ 4.73)& 80.74 ($\uparrow$ 5.18)& 80.52 ($\uparrow$ 5.56)& 77.24 ($\uparrow$ 0.18)& 73.92 ($\uparrow$ 0.22)& 75.05 ($\uparrow$ 0.25)\\
    3\% & baseline& 55.41 & 59.47 & 56.41 & 68.98 & 69.88 & 68.7& 49.43 & 59.17 & 51.77\\
    3\% & DF-ALC (w/RL)& 56.67 ($\uparrow$ 1.26)& 60.87 ($\uparrow$ 1.4)& 57.51 ($\uparrow$ 1.1)& 72.67($\uparrow$ 3.69) & 73.35 ($\uparrow$ 3.47)& 71.84 ($\uparrow$ 3.14)& 50.2 ($\uparrow$ 0.77)& 60.79 ($\uparrow$ 1.62)& 52.73 ($\uparrow$ 0.96)\\
    5\% & baseline& 46.87 & 51.91 & 48.37 & 63.68 & 65.52 & 64.03& 42.73& 52.40& 44.29\\
    5\% & DF-ALC (w/RL)& 48.45 ($\uparrow$ 1.58)& 53.93 ($\uparrow$ 2.02)& 49.91 ($\uparrow$ 1.54)& 67.06 ($\uparrow$ 3.38)& 68.56 ($\uparrow$ 3.04)& 66.9 ($\uparrow$ 2.87)& 43.45 ($\uparrow$ 0.72)& 54.05 ($\uparrow$ 1.65)& 45.18 ($\uparrow$ 0.89)\\
    10\% & baseline& 36.23 & 41.21 & 37.31 & 55.7 & 58.74 & 56.69& 32.94& 40.25& 32.68\\
    10\% & DF-ALC (w/RL)& 37.35 ($\uparrow$ 1.12)& 42.71 ($\uparrow$ 1.5)& 38.45 ($\uparrow$ 1.14)& 58.82 ($\uparrow$ 3.12)& 61.71 ($\uparrow$ 2.97)& 59.5 ($\uparrow$ 2.81)& 33.53 ($\uparrow$ 0.59)& 41.66 ($\uparrow$ 1.41)& 33.35 ($\uparrow$ 0.67)\\

   \bottomrule

\end{tabular} }
\caption{Low-resource image object classification results. Mask rate denotes the rate of features extracted by FRCNN (baseline) that are randomly assigned.}
 \label{tab:lowresourse}
\end{table*}




\section{Conclusion and Future Work}
\label{sec:conclusion}

In this work, we have presented DF-$\mathcal{ALC}$, a differentiable fuzzy description logic language for symbol grounding, which is also the first representation learning method for $\mathcal{ALC}$ ontologies. We have proved the soundness of the semantics of DF-$\mathcal{ALC}$ under OWA and the soundness of learning to ground. And we pointed out the limitations of directly using the semantics of fuzzy $\mathcal{ALC}$ in DF-$\mathcal{ALC}$, so we also presented a rule-based loss for symbol grounding, which is effective in the semantic image interpretation task and CQs answering task. If a given $\mathcal{ALC}$ ontology is complete, directed by a perceptual neural network, learning in the transformed DF-$\mathcal{ALC}$ ontology can learn an interpretable grounding the crisp transformed of which is the model the $\mathcal{ALC}$ ontology and can remain much valuable information from the perceptual neural network. Compared with the most related differentiable fuzzy logic model, LTN, we find that DF-$\mathcal{ALC}$ is better at retaining the reliable part of probability. Besides, DF-$\mathcal{ALC}$ is under the open-world assumption (OWA) which is more robust and close to realistic situations. 

However, current DF-$\mathcal{ALC}$ does not do well in part-of relation recognition revision, which remains to be tested and updated for other link prediction tasks. Besides, $\mathcal{ALC}$ cannot express cardinality constraints, which is important in situations where the definition of concepts is highly sensitive to quantity. So to extend expressive power, we will explore the best differentiable fuzzy model for $\mathcal{ALCN}$. Our next step is to construct commonsense knowledge in $\mathcal{ALC}$ and apply DF-$\mathcal{ALC}$ in combining knowledge into a dialogue state tracker. 

\bibliography{aaai22}

\bigskip

\end{document}